\documentclass[10pt,twocolumn,letterpaper]{article}

\usepackage{cvpr}              

\definecolor{cvprblue}{rgb}{0.21,0.49,0.74}
\usepackage[pagebackref,breaklinks,colorlinks,allcolors=cvprblue]{hyperref}
\usepackage{placeins}

\def\paperID{9} 
\def\confName{DICTA}
\def\confYear{2025}

\title{Filling the Gaps: A Multitask Hybrid Multiscale Generative Framework for\\ 
Missing Modality in Remote Sensing Semantic Segmentation}

\author{Nhi~Kieu \and Kien~Nguyen \and Arnold~Wiliem \and Clinton~Fookes \and Sridha~Sridharan\\
School of Electrical Engineering and Robotics, Queensland University of Technology\\
Brisbane, QLD, Australia\\
{\tt\small \{v.kieu,nguyentk,a2.wiliem,c.fookes,s.sridharan\}@qut.edu.au}
\and
Arnold Wiliem\\
Shield AI\\
{\tt\small arnold.wiliem@shield.ai}
}

\begin{document}
\maketitle
\begin{abstract}
Multimodal learning has shown significant performance boost compared to ordinary unimodal models across various domains. However, in real-world scenarios, multimodal signals are susceptible to missing because of sensor failures and adverse weather conditions, which drastically deteriorates models' operation and performance. Generative models such as AutoEncoder (AE) and Generative Adversarial Network (GAN) are intuitive solutions aiming to reconstruct missing modality from available ones. Yet, their efficacy in remote sensing semantic segmentation remains underexplored. In this paper, we first examine the limitations of existing generative approaches in handling the heterogeneity of multimodal remote sensing data. They inadequately capture semantic context in complex scenes with large intra-class and small inter-class variation. In addition, traditional generative models are susceptible to heavy dependence on the dominant modality, introducing bias that affects model robustness under missing modality conditions. To tackle these limitations, we propose a novel \textbf{G}enerative-\textbf{E}nhanced \textbf{M}ulti\textbf{M}odal learning \textbf{Net}work (GEMMNet) with three key components: (1) Hybrid Feature Extractor (HyFEx) to effectively learn modality-specific representations, (2) Hybrid Fusion with Multiscale Awareness (HyFMA) to capture modality-synergistic semantic context across scales and (3) Complementary Loss (CoLoss) scheme to alleviate the inherent bias by encouraging consistency across modalities and tasks. Our method, GEMMNet, outperforms both generative baselines AE, cGAN (conditional GAN), and state-of-the-art non-generative approaches - mmformer and shaspec - on two challenging semantic segmentation remote sensing datasets (Vaihingen and Potsdam). Source code is made available {here}\footnote{https://github.com/nhikieu/GEMMNet}.
\end{abstract}    
\section{Introduction}
Multimodal learning has gained popularity in machine learning applications, as it significantly enhances model performance compared to traditional unimodal networks \cite{zhang2021deep, tawfik2021survey, zhu2024review}. Yet, in practical settings, multimodal signals are prone to missing due to complicated data acquisition processes, unstructured data sources, and device failures \cite{Rahate2022203}. A substantial research effort has been invested in improving model robustness towards missing modality scenarios across domains, such as medical imaging \cite{zhou2023literature} and vision-language-audio \cite{ma2022multimodal}. 

\begin{figure}[t]
  \centering
  \includegraphics[width=\linewidth]{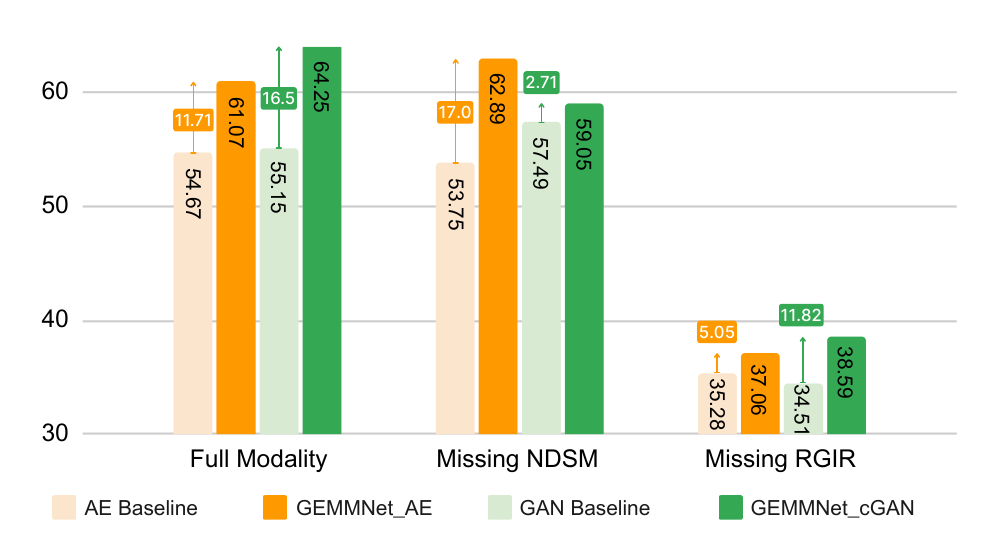}
  \caption{Highlights of F1 score on a challenging "car" class on the Vaihingen dataset (performance improvement shown in relative percentage). Our GEMMNet significantly boosts the performance of both generative models AE and cGAN across scenarios: Full Modality, Missing NDSM, Missing RGIR.}
  \label{fig:bar_chart_vaihingen}
\end{figure}

\begin{figure}[t]
  \centering
  \includegraphics[width=\linewidth]{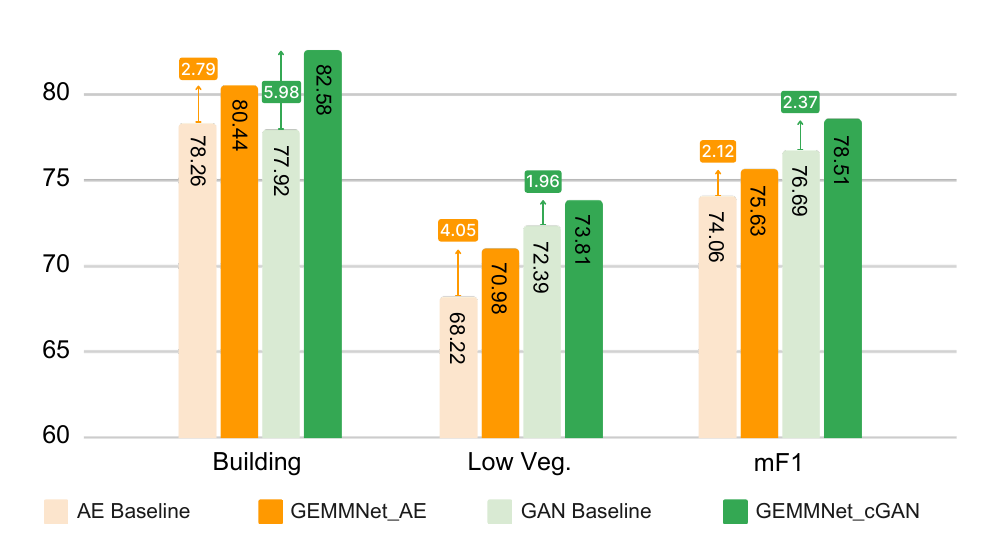}
  \caption{Highlights of F1 score on the Potsdam dataset when missing NDSM modality (performance improvement shown in relative percentage). Our GEMMNet demonstrates noticeable gain on classes such as building and low vegetation and overall segmentation (mean F1) compared to ordinary generative models.}
  \label{fig:bar_chart_potsdam}
\end{figure}

\vspace{4px}
 \noindent Although remote sensing (RS) applications are also likely to encounter missing modalities during inference due to sensor malfunctions and harsh operating environments, this area remains underexplored. Generative models, particularly Generative Adversarial Network (GAN) and AutoEncoder (AE), have been widely employed in other domains to address missing modalities \cite{zhou2023literature, zhou2020hi, wang2023incomplete}. Such approaches explicitly reconstruct missing modalities, providing intuitive implementation demanding little to no modification to existing systems with great interpretability. Their self-supervised learning capability enables effective feature extraction even under severely missing modality conditions, as demonstrated by a recent study \cite{ma2021smil}.  However, existing works in remote sensing overlook the potential of generative models, which have been preliminarily studied in the early days for binary segmentation and classification tasks \cite{bischke2018overcoming, pande2019adversarial}. The power of generative models in the semantic segmentation task, with the incorporation of other advancements such as attention mechanisms and multitask learning, has been neglected in remote sensing, which has been proven effective in other domains \cite{zhang2022mmformer, wang2023multi}. To address the gap, we propose a novel GEMMNet (\textbf{G}enerative-\textbf{E}nhanced \textbf{M}ulti\textbf{M}odal learning \textbf{Net}work), which integrates hybrid feature extraction, hybrid multiscale fusion and multitask learning. The proposed method leverages the strengths of generative models (cGAN and AE) to handle missing modality scenarios in the remote sensing semantic segmentation task. 
 
 \vspace{4px}
 \noindent Our proposed network, GEMMNet, introduces three key innovations. A HyFEx module incorporates convolution and transformer blocks to capture local and global features within the missing modality generator and modality-specific encoder. HyFMA is a dynamic fusion module that integrates features across scales and modalities through a hybrid attention-convolution mechanism. CoLoss is a multi-objective loss that combines modality-specific supervision and multiscale fusion consistency, encouraging the model to learn both discriminative and generalizable representations.  \Cref{fig:bar_chart_vaihingen} and \Cref{fig:bar_chart_potsdam} highlight the superior performance of our method over baseline generative models (AE and cGAN) on challenging classes on two complex remote sensing datasets: Vaihingen and Potsdam. On the other hand, our GEMMNet also surpasses state-of-the-art methods using other learning regimes adopted from other domains (mmformer \cite{zhang2022mmformer} and shaspec \cite{wang2023multi}) 

\vspace{4px}
\noindent The contributions of this paper are fourfold:
\begin{itemize}
    \item To our knowledge, our work is the first to assess the robustness of generative models on missing modality for the RS semantic segmentation task.
    \item We introduce a novel \textbf{G}enerative-\textbf{E}nhanced \textbf{M}ulti\textbf{M}odal learning \textbf{Net}work (GEMMNet) consisting of three novel components (HyFEx, HyFMA and CoLoss) designed to facilitate semantic context learning and multitask learning to improve the robustness of generative models against missing modality in RS.
    \item Extensive experiments on widely benchmarked RS datasets - Vaihingen, Potsdam - validate the superior performance of our method over baseline generative models and state-of-the-art methods from other learning regimes.
    \item The source code and pretrained models will be made publicly accessible to promote transparency and facilitate further research in this area.  
\end{itemize}
\section{Related Works}
\textbf{Multimodal Learning.} The determining component of a powerful multimodal learning method is multimodal features fusion. Modality-specific and modality-synergistic features need to be effectively captured to produce an accurate prediction. Multimodal fusion methods can be categorized into three main categories: Early fusion, Intermediate fusion, and Late fusion \cite{kieu2024multimodal, zhang2021deep}. Early fusion approaches \cite{kieu2023general, tang2022matr} are simpler because of no need for unimodal separate processing branch; however, it requires modalities to be aligned at the input level and is vulnerable to missing modality. Whereas, methods applying late fusion \cite{kampffmeyer2018urban, zhang2024multimodal, fan2023pmr} allow more flexible incorporation of features from different modalities at the cost of limited multimodal synergy. Intermediate fusion \cite{hong2020more, li2022dense, kumar2021improved} often strikes the balance between the two aforementioned schemes, where modality-specific features from different modalities are fused to create unified multimodal representations for further learning down the architecture pipeline. To best leverage the advantages of each fusion strategy, a hybrid approach combining multi-level fusion schemes can be utilized \cite{tawfik2021survey, zhu2024review, li2022dense}. \textit{Our work proposes HyFMA - Hybrid Fusion with Multiscale Awareness - to effectively capture modality-specific and modality-synergistic features using attention and convolution mechanisms. Multiscale learning is beneficial for semantic segmentation \cite{tang2022matr, zhou2020hi, zhang2022mmformer}, especially for dealing with huge scale variation in remote sensing data \cite{kieu2023general}.}
\\
\\
\noindent\textbf{Missing Modality.} Multimodal learning is susceptible to significant performance degradation due to missing modality because of sensor failures and unfavorable operating conditions in deployment \cite{Rahate2022203, zhou2023literature}. Generative models such as AutoEncoder (AE) and Generative Adversarial Network (GAN) are the most intuitive methods yet among the most powerful ones by explicitly reconstructing missing modality from available inputs. Their efficacy has been highlighted in various domains such as Medical Imaging \cite{zhou2023literature, zhou2020hi, cao2020auto, huang2021aw3m} and Vision-Language-Audio \cite{wang2023incomplete, luo2023multimodal, ma2021smil}. However, as pointed out in the aforementioned papers, GAN-based and AE models need to be further enhanced to adapt to data nature. Adding auxiliary loss terms can foster richer feature learning \cite{zhou2020hi, zhang2022mmformer, cao2020auto, huang2021aw3m}. Therefore, \textit{we design the CoLoss objective function to facilitate robust multimodal learning in remote sensing to deal with missing signals. On the other hand, joint usage of convolution and attention mechanisms also allows more effective feature extraction \cite{zhou2020hi, zhang2022mmformer}, which motivates us to integrate HyFEx into our pipeline. Moreover, attention mechanism can act as a dynamic modality weighting scheme to deal with missing modality \cite{zhang2022mmformer, wang2023incomplete}. Our HyFMA jointly leverages attention and convolution operations to enhance the robustness of multimodal fusion towards missing modality.}
\\
\\
\noindent\textbf{Missing Modality in Remote Sensing}. Despite significant efforts in tackling missing modality in other domains, remote sensing remains underexplored. A recent survey by \cite{kieu2024multimodal} emphasizes the need for further research on handling missing modality within remote sensing applications. Existing studies addressing missing modalities in remote sensing span across classification \cite{pande2019adversarial, kumar2021improved, wei2023msh} and semantic segmentation \cite{kampffmeyer2018urban, bischke2018overcoming, kang2022disoptnet, li2022dense, chen2024novel}. Early works \cite{bischke2018overcoming, pande2019adversarial} demonstrate promising results of generative models in handling missing modalities in remote sensing on binary segmentation and classification tasks, respectively. However, recent studies have overlooked the potential of generative models for the semantic segmentation task with the incorporation of an attention mechanism. Such explicit synthesis approaches not only offer great interpretability but also inherently support unsupervised learning, enabling effective training even when labeled data is scarce or incomplete. Particularly, when modalities are frequently missing, generative models can effectively leverage the available unpaired data. Modern approaches such as SMIL \cite{ma2021smil} have shown that utilizing generative models' strengths can significantly enhance model robustness against severe modality scarcity. \textit{However, current generative approaches in remote sensing have neglected the power of hybrid multiscale learning and multitask learning, despite their proven effectiveness in other domains as discussed previously. Our approach addresses this gap by proposing a novel GEMMNet architecture integrating a multitask hybrid multiscale fusion strategy.} Through its three key innovations (i.e., HyFEx, HyFMA, and CoLoss), the proposed method significantly improves the robustness of generative models towards missing modality in remote sensing semantic segmentation. Our GEMMNet surpasses other state-of-the-art methods (mmformer and shaspec) from other learning regimes adopted from different domains.
\section{Methodology}
\begin{figure*}
  \centering
  \includegraphics[width=0.8\textwidth]{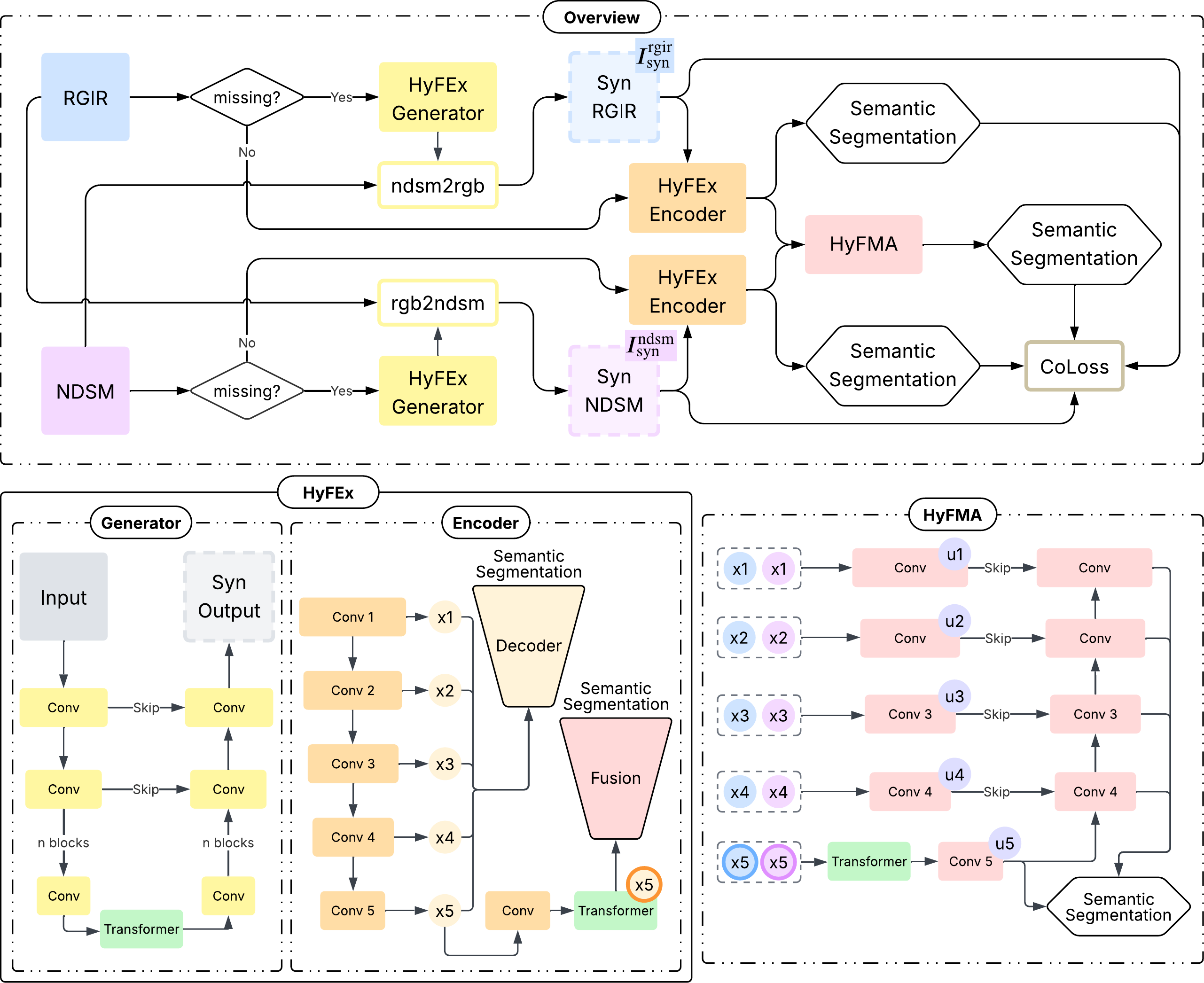}
  \caption{Overview of our proposed method GEMMNet comprising three main components: Hybrid Feature Extractor (HyFEx), Hybrid Fusion with Multiscale Awareness (HyFMA), and Complementary Loss terms (CoLoss).}
  \label{fig:model_schema}
\end{figure*}

Our proposed framework enhances the robustness of generative models towards missing modalities by purposefully designing three key novel components to address the challenges in remote sensing data: (1) Hybrid Feature Extractor (HyFEx), (2) Hybrid Fusion with Multiscale Awareness (HyFMA), and (3) Complementary Loss Scheme (CoLoss). 

\vspace{6px}
\noindent\Cref{fig:model_schema} illustrates the workflow of the proposed method. The model receives an image from each modality $\boldsymbol{I}^m \in \mathbb{R}^{C\times W \times H}$ where $m$ denotes modality type $m \in \{\text{rgir, ndsm}\}$. The HyFEx Generator synthesizes missing modality signals $\boldsymbol{I}^m_{\text{syn}} \in \mathbb{R}^{C\times W \times H}$ from the available one. Three scenarios are uniformly randomized in training by a modality mask $\in \{\text{[True, True], [True, False], [False, True]}\}$ for full modality, missing ndsm and missing rgir, respectively. Based on which, either real $\boldsymbol{I}^m$ or synthesized $\boldsymbol{I}^m_{\text{syn}}$ input signals are processed by the HyFEx Encoder into unimodal multiscale features $\{\boldsymbol{x}^m_i\}^5_{i=1}, \boldsymbol{x}^m_i \in \mathbb{R}^{C_i \times W_i \times H_i}$, where $5$ is the number of pyramid level. These modality-specific features $\{\boldsymbol{x}^m_i\}^5_{i=1}$ are then fused into unified multiscale representations $\{\boldsymbol{u}_i\}^5_{i=1}, \boldsymbol{u}_i \in \mathbb{R}^{C_i \times W_i \times H_i}$ via HyFMA and decoded to the final semantic segmentation map. Finally, CoLoss integrates multiple loss terms to boost model robustness against missing modality scenarios.



\subsection{Hybrid Feature Extractor (HyFEx)}
Our HyFEx employs a hybrid convolution and transformer \cite{vaswani2017attention} architecture for both the missing modality Generator and unimodal Encoder modules. Conventional convolutional architectures of AE \cite{bank2023autoencoders} and cGAN (conditional GAN) Pix2Pix \cite{isola2017image} 
are insufficient for capturing long-range spatial dependencies \cite{nyamathulla2024analysis}, limiting their contextual understanding of complex scenes. The hybrid convolution and transformer enables better modeling of global and local context \cite{zhang_jiang_2022}. Specifically, convolutional layers extract local spatial features in the early stages, while transformer modules effectively model global context at bottleneck layers. The Generator synthesizes missing modality creating $\boldsymbol{I}^m_{\text{syn}}$:
\begin{equation}
    \boldsymbol{I}^m_{\text{syn}} = \operatorname{G}(\boldsymbol{I}^{\text{avail}};\theta_{\operatorname{G}}),
\end{equation}
where $\operatorname{G}(\cdot)$ represents the Generator parameterized by $\theta_{\operatorname{G}}$ and $\boldsymbol{I}^{\text{avail}}$ indicates the available modality. HyFEx Encoder $\operatorname{E}(\cdot)$ is composed of convolutional layers and a transformer block at bottleneck parameterized by $\theta_{\operatorname{E}}$ extracting feature for each modality input $\boldsymbol{I}^m$:
\begin{equation}
    \{\boldsymbol{x}^m_i\}^5_{i=1} = \operatorname{E}(\boldsymbol{I}^m; \theta_{\operatorname{E}}).
\end{equation}

\subsection{Hybrid Fusion with Multiscale Awareness (HyFMA)}
Accurately integrating multimodal data demands consideration of multiscale spatial contexts, as features across different resolutions capture complementary semantic and geometric information. To address this, we introduce HyFMA, a fusion mechanism explicitly designed for multiscale feature integration. Rather than simple concatenation or summation, HyFMA leverages both convolution and transformer to dynamically weighs modality-specific features according to their relevance at various scales. For scales $i=[1,4]$ fusion is performed via convolution:
\begin{equation}
    \{\boldsymbol{u}_i\}^4_{i=1} = \operatorname{Conv}(\operatorname{Concat}(\boldsymbol{x}_i^A, \boldsymbol{x}_i^B); \theta^{(i)}_{\operatorname{Conv}}),
\end{equation}
where $A, B$ are different modalities and $\operatorname{Conv}(\cdot)$ denotes series of convolution layers parameterized by $\theta^{(i)}_{\operatorname{Conv}}$. For the bottleneck layer $i=5$, the transformer attention mechanism $\operatorname{Trans}(\cdot)$ is applied to fuse concatenated features into a unified representation to model global dependencies (i.e., semantic context of the scene) since features at deeper level are more abstract and high level. On the other hand, convolutional operation is used as fusion mechanism in earlier layers $i \in [1, 4]$ to effectively capture local features at lower level.
\begin{equation}
    u_5=\operatorname{Trans}(\operatorname{Concat}(\boldsymbol{x}_5^A, \boldsymbol{x}_5^B); \theta_{\operatorname{Trans}}),
\end{equation}
where $\theta_{\operatorname{Trans}}$ are transformer module parameters. These multiscale unified representations $\{\boldsymbol{u}_i\}^5_{i=1}$ are decoded to semantic segmentation maps through a Decoder $\operatorname{D}(\cdot)$. We designed the Decoder following a UNet-like \cite{ronneberger2015u} manner. The final semantic segmentation map $\boldsymbol{S} \in \mathbb{R}^{C\times H \times W}$ is:
\begin{equation}
    \boldsymbol{S}=\operatorname{D}(\{\boldsymbol{u}_i\}^5_{i=1};\theta_{\operatorname{D}}),
\end{equation}
where $C$ is the number of classes and $\theta_{\operatorname{D}}$ are parameters of the Decoder $\operatorname{D}(\cdot)$.

\subsection{Complementary Loss (CoLoss) scheme}
The performance and robustness of multimodal segmentation networks can significantly benefit from leveraging auxiliary learning tasks. We design CoLoss, which is a multi-objective loss that combines missing modality reconstruction, modality-specific supervision and multiscale fusion consistency, encouraging the model to learn both discriminative and generalizable representations. Thus, it promotes robust feature learning in missing modality settings. CoLoss incorporates four main loss terms:
\begin{equation}
    \mathcal{L}_{\text{CoLoss}} = \mathcal{L}_{seg}^{\text{fused}} + \mathcal{L}_{seg}^{\text{fused\_scales}} + \mathcal{L}_{seg}^m + \mathcal{L}_{rec},
\end{equation}
where $\mathcal{L}_{seg}^{\text{fused}}$ is a semantic segmentation loss calculated using prediction $\boldsymbol{S}$ by aggregating all fused features $\{\boldsymbol{u}_i\}^5_{i=1}$ through the Decoder $\operatorname{D}(\cdot)$. On the other hand, $\mathcal{L}_{seg}^{\text{fused\_scales}}$ is a summation of semantic segmentation losses calculated at four different scales using unified features at each level $\{\boldsymbol{u}_i\}^4_{i=1}$. $\mathcal{L}_{seg}^m$ is a semantic segmentation loss of each modality $m$ (i.e., unimodal semantic segmentation loss). 

Both datasets are heavily imbalanced (i.e., car class occupies just over 1\%), hence we employ a joint loss of $\operatorname{Dice}(\cdot)$ and Weighted Soft Cross Entropy $\operatorname{Weighted\_CE}(\cdot)$ \cite{kieu2023general} for each semantic segmentation prediction: 
\begin{equation}
    \mathcal{L}_{seg}=\operatorname{Dice}(y, \hat{y}) + \operatorname{Weighted\_CE}(y, \hat{y}),
\end{equation}
where $y$ is the true semantic segmentation map (i.e., label) and $\hat{y}$ is the predicted one.

Depending on the missing modality reconstruction algorithm, which is either AE \cite{bank2023autoencoders} or cGAN Pix2Pix \cite{isola2017image},  reconstruction loss $\mathcal{L}_{rec}$ will be formulated differently as follows:
\begin{equation}
    \mathcal{L}^{AE}_{rec} = ||\boldsymbol{I}^m-\boldsymbol{I}^m_{\text{syn}}||_2^2,
\end{equation}
\begin{equation}
    \mathcal{L}^{cGAN}_{rec} = \arg \min_{\operatorname{G}} \max_{\operatorname{D}} \; \mathcal{L}_{cGAN}(\operatorname{G}, \operatorname{D}) + \lambda \mathcal{L}_{L_1}(\operatorname{G}),
\end{equation}
where the AE reconstruction loss $\mathcal{L}^{AE}_{rec}$ uses a L2 norm. On the other hand, $\mathcal{L}_{cGAN}(\operatorname{G}, \operatorname{D})$ aims to optimize reconstruction through a competitive process between two networks: a generator $\operatorname{G}(\cdot)$ and a discriminator $\operatorname{D}(\cdot)$. In addition, a L1 loss, $\mathcal{L}_{L_1}(\operatorname{G})$, is included to encourage the generator $\operatorname{G}$ to produce more realistic synthesized signals.
\section{Experiments}
\begin{table*}[t]
  \centering
  \caption{Evaluation results on Vaihingen and Potsdam datasets. Class-wise F1 scores are recoded with differences between baseline and GEMMNet models in relative percentage for AE and cGAN ($\Delta$). Significant performance gain is highlighted in green. Models mmformer \cite{zhang2022mmformer} and shaspec \cite{wang2023multi} are our implementations adapted from other domains to remote sensing data.}
  \label{tab:vaihingen_potsdam}
  \includegraphics[width=\textwidth]{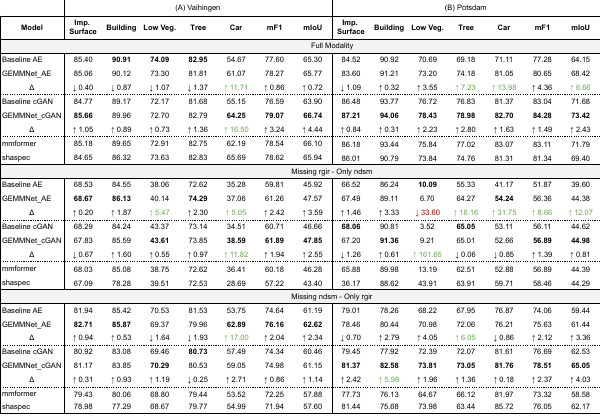}
\end{table*}

\subsection{Experimental Setting}
We conduct our experiments on two popular datasets in remote sensing semantic segmentation.
\textbf{Vaihingen \cite{Vaihingen}:} This dataset from the International Society for Photogrammetry and Remote Sensing (ISPRS) contains remote sensing data of the Vaihingen region in Germany. It has two modalities: RGIR and NDSM. It contains 33 large image tiles of different sizes with a GSD of 9 cm.
\textbf{Potsdam \cite{Potsdam}:} This dataset, also from the ISPRS, contains remote sensing data of the Potsdam region in Germany. The dataset contains 38 patches of the same size, each consisting of an RGIR and an NDSM. The ground sampling distance of both is 5 cm.

\vspace{3px}
\noindent Selected tiles for train, validation, and test are as specified on the ISPRS data portal. Samples are extracted from both datasets with a size of 512$\times$512 and random augmentation. The models presented in this paper were trained on an NVIDIA GeForce RTX 3090 GPU. We evaluate model performance using F1 score per class, mean F1 score (mF1), and mean Intersection over Union (mIoU).

\begin{figure*}
  \centering
  \includegraphics[width=0.96\textwidth]{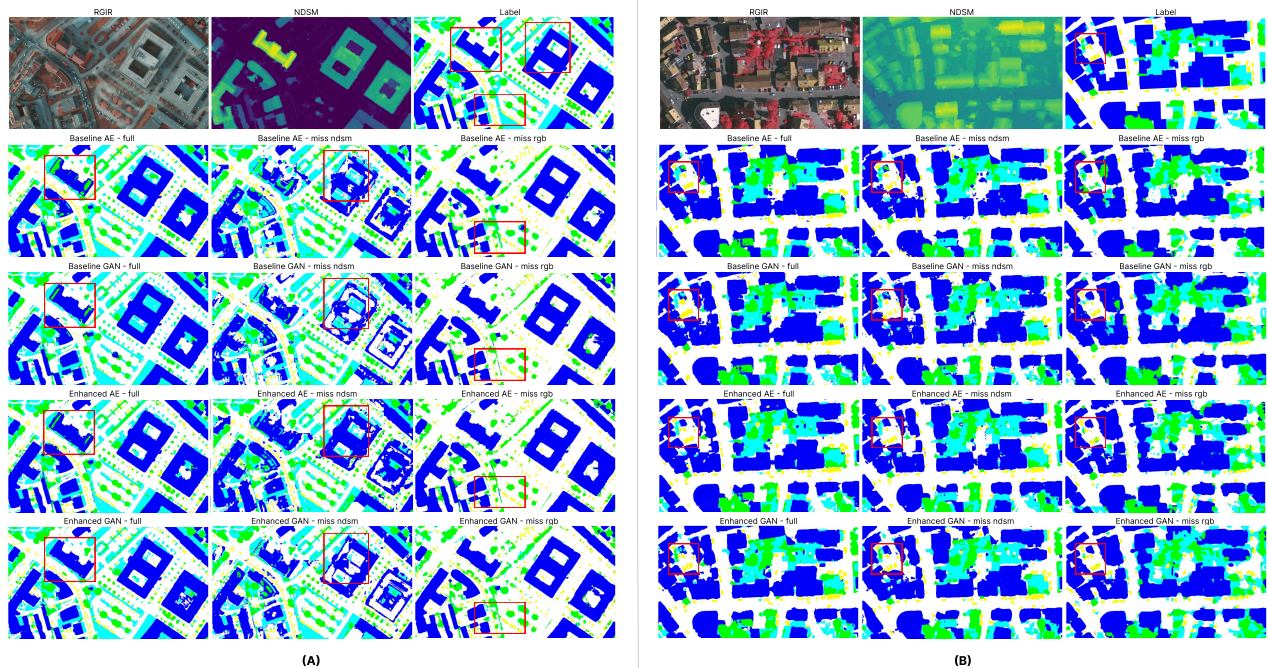}
  \caption{Qualitative Results on (A) Potsdam and (B) Vaihingen. Our methods produce more accurate and realistic boundaries in all three scenarios - full modality, missing RGB and missing NDSM (highlighted in red boxes).}
  \label{fig:qualitative}
\end{figure*}

\subsection{Result Analysis}
To comprehensively evaluate the robustness of our proposed enhancements towards missing modality in remote sensing semantic segmentation, we present experimental results across three scenarios on two datasets: (1) full modality available, (2) missing rgir (only ndsm available), and (3) missing ndsm (only rgir available).

\noindent\textbf{Quantitative Analysis:} \Cref{tab:vaihingen_potsdam} reports per-class F1 scores as well as mean F1 (mF1) and mean Intersection-over-Union (mIoU) for all models. Our GEMMNet models applied to both AE and cGAN consistently outperform their respective baselines on both datasets. Performance gain ($\Delta$) is shown in relative percentages, unless stated otherwise.

\begin{itemize}
  \item \textit{Full Modality.} The overall performance gain on the Vaihingen benchmark was modest for the AE baseline (mF1 rose from 76.59 to 78.27), whereas the cGAN-based counterpart exhibited a larger uplift (mF1 improved from 77.60 to 79.07). Crucially, our enhancements in GEMMNet delivered pronounced benefits for the challenging ‘car’ class. Specifically, the class-wise F1 score for the AE model climbed from 54.67 to 61.07 (+11.7\%), while that of the GAN-based model increased from 55.15 to 64.25 (+16.5\%). These findings demonstrate that integrating a transformer module for richer contextual encoding, coupled with multiscale awareness through our HyFMA and CoLoss, effectively boost semantic segmentation performance, especially in difficult classes.
  \item \textit{Missing RGIR - Only NDSM Available.} In the more challenging scenario where the RGIR modality is missing (only NDSM is available), performance drops substantially across all models, confirming the significant role of spectral information in segmentation tasks. However, our GEMMNet models enhancing both AE and cGAN baselines still deliver consistent gains. On the Vaihingen, cGAN-based model observed a slight overall improvement from 60.71 to 61.89 and from 46.66 to 47.85 for mean F1 and mean IoU, respectively. However, the upturn for the car class was remarkable, as seen in the full modality case, from 34.51 to 38.59 (+11.82\%). A similar trend is realized for AE-based models. On the Potsdam dataset, our GEMMNet model applied to AE experienced a gain of 31.7\% on ‘car' class (from 41.17 to 54.24); while the strengthened GAN-based model observed a 2.6-fold increase on the ‘low-vegetation' category (3.52 $\rightarrow$ 9.21). This suggests that the transformer component effectively lessens the dependence on a strong prior of NDSM. Although the low vegetation (low veg.) shows a relative F1 drop of 33.6\% under the missing rgir scenario on Potsdam dataset, this figure exaggerated the impact. The low veg. F1 score of the baseline AE model is already very low ($\sim$10\%). Therefore a modest absolute decline of $\sim$3.4 percentage points (pp) translates into a large relative decrease. In fact, this minor adjustment allows our GEMMNet model to prioritize more high-return classes - improving tree by +8.94 pp and car by +13.07 pp - thereby maximizing overall robustness.
  \item \textit{Missing NDSM - Only RGIR Available.} When only the RGIR modality is present, overall performance surpassed that of the previous scenario, confirming that the RGIR modality contains richer information for most land cover classes. Once again, our GEMMNet models delivered consistent performance gains on both Vaihingen and Potsdam across all classes with some clear winners. On Potsdam, our GEMMNet model applied to cGAN has an almost 6\% increase for ‘building' class (77.92 $\rightarrow$ 82.58). Meanwhile, the strengthened AE model achieves over 6\% increase for ‘tree' class (67.95 $\rightarrow$ 72.06). On Vaihingen, the dominant gainer remains the ‘car' class for the AE model, whose F1 surges by 17\% (53.75 $\rightarrow$ 62.89).
  \item \textit{Compare with other learning regimes.} For completeness, we extended the application of mmformer \cite{zhang2022mmformer} and shaspec \cite{wang2023multi} to remote sensing data. They are state-of-the-art methods dealing with missing modality in other domains using different learning regimes. As can be seen from \Cref{tab:vaihingen_potsdam}, our proposed GEMMNet models surpass both mmformer and shaspec in terms of mF1 and mIoU across all scenarios: full modality, missing rgir and missing ndsm.
\end{itemize}

\noindent \textbf{Qualitative Analysis:}  As illustrated in \Cref{fig:qualitative}, our proposed methods consistently outperform the baseline AE and cGAN models, especially on a challenging class like ‘car', which exhibits huge scale variation and is extremely scarce. Our predicted class boundaries are noticeably sharper and smoother, creating a more realistic semantic segmentation map. This uplift can be attributed to the multiscale-aware mechanism in the fusion component HyFMA and multitask optimizing CoLoss, which ensures the retention of both coarse- and fine-grained information. Thereby, it greatly enhances model robustness, especially under missing modality scenarios. However, our ablative experiments revealed diminishing returns and significantly higher computational costs when transformers were employed across multiple feature levels; thus, we restricted transformer usage exclusively to bottleneck representations. This hybrid approach strikes an effective balance between context-awareness and computational efficiency.

\vspace{3px}
\noindent Interestingly, in the full modality use case on Potsdam, roofs that have a very similar color to ‘tree' (highlighted in red boxes on the first column of \Cref{fig:qualitative} - A), are wrongly classified as ‘tree' instead of ‘building'. Our proposed method minimized such error on the AE model and totally eliminated the issue on the cGAN model. This is evident that adding a transformer can effectively capture contextual information to understand complex scenes.
\section{Conclusion}
In this paper, we introduced a novel \textbf{G}enerative-\textbf{E}nhanced \textbf{M}ulti\textbf{M}odal learning \textbf{Net}work (GEMMNet) that remains robust under arbitrary missing modalities scenarios targeting remote sensing semantic segmentation. Our framework is composed of three main innovations: Hybrid Feature Extractor (HyFEx) for missing modality reconstruction and modality-specific encoding, Hybrid Fusion with Multiscale Awareness (HyFMA) module for context-adaptive multimodal integration, and a Complementary Loss (CoLoss) scheme that fosters rich feature learning. Our GEMMNet models consistently outperformed AE and cGAN baselines as well as two other powerful frameworks from other domains (mmformer and shaspec) on the Vaihingen and Potsdam benchmarks. Gains were most pronounced for challenging, scale-varying classes - e.g., up to +17\% and +31.75\% F1 for 'car' on Vaihingen and Potsdam, respectively, in the missing RGIR modality case. These results offer a practical route to resilient remote sensing multimodal learning systems. 

\section*{Acknowledgment}
This work partly supported by Shield AI. Shield AI is one of the global leaders in developing AI pilots for defence and civilian applications.
\FloatBarrier
{
    \small
    \bibliographystyle{ieeenat_fullname}
    \bibliography{main}
}
\end{document}